%% file: main.tex
\PassOptionsToPackage{numbers}{natbib}

\documentclass{article}
\usepackage[ruled,vlined,linesnumbered]{algorithm2e}

\usepackage[final]{neurips_2024}

\usepackage[utf8]{inputenc} 
\usepackage[T1]{fontenc}    
\usepackage{hyperref}       
\usepackage{url}            
\usepackage{booktabs}       
\usepackage{amsfonts}       
\usepackage{nicefrac}       
\usepackage{microtype}      
\usepackage{xcolor}         

\usepackage{caption}
\usepackage{subcaption}
\usepackage{graphicx} 
\usepackage{color}
\usepackage{graphicx}
\usepackage{amsmath}
\usepackage{cleveref}
\usepackage{makecell}
\usepackage{etoolbox}
\usepackage{enumitem}
\usepackage{wrapfig}

\usepackage{float}
\usepackage{xfrac}

\usepackage{tikz}

\makeatletter
\renewcommand{\SetKwInOut}[2]{%
  \sbox\algocf@inoutbox{\KwSty{#2}\algocf@typo:}%
  \expandafter\ifx\csname InOutSizeDefined\endcsname\relax
    \newcommand\InOutSizeDefined{}\setlength{\inoutsize}{\wd\algocf@inoutbox}%
    \sbox\algocf@inoutbox{\parbox[t]{\inoutsize}{\KwSty{#2}\algocf@typo:\hfill}~}\setlength{\inoutindent}{\wd\algocf@inoutbox}%
  \else
    \ifdim\wd\algocf@inoutbox>\inoutsize%
    \setlength{\inoutsize}{\wd\algocf@inoutbox}%
    \sbox\algocf@inoutbox{\parbox[t]{\inoutsize}{\KwSty{#2}\algocf@typo:\hfill}~}\setlength{\inoutindent}{\wd\algocf@inoutbox}%
    \fi%
  \fi
  \algocf@newcommand{#1}[1]{%
    \ifthenelse{\boolean{algocf@inoutnumbered}}{\relax}{\everypar={\relax}}%
    {\let\\\algocf@newinout\hangindent=\inoutindent\hangafter=1\parbox[t]{\inoutsize}{\KwSty{#2}\algocf@typo:\hfill}~##1\par}%
    \algocf@linesnumbered
  }}%
\makeatother

\SetKwInput{KwInput}{Input}
\SetKwInput{KwOutput}{Output}

\definecolor{darkgreen}{rgb}{0.0, 0.5, 0.0}
\newcommand{\red}[1]{\color{red}#1}
\newcommand{\green}[1]{\color{darkgreen}#1}

\newcommand{\algname}[0]{MESS+\xspace}

\title{
    \algname: Energy-Optimal Inferencing in Language Model Zoos with Service Level Guarantees 
}

\author{%
Ryan Zhang \\
Horace Greeley High School\\
\texttt{ryzhangofficial@gmail.com} 
\And
Herbert Woisetschl\"ager \\
Technical University of Munich\\
\texttt{h.woisetschlaeger@tum.de} 
\AND
\hspace{1.1em}Shiqiang Wang \\
\hspace{1.1em}IBM Research\\
\hspace{1.1em}\texttt{wangshiq@us.ibm.com} 
\And
\hspace{0.5em}Hans-Arno Jacobsen \\
\hspace{0.5em}University of Toronto\\
\hspace{0.5em}\texttt{jacobsen@eecg.toronto.edu} 
}

\begin{document}

\maketitle

\begin{abstract}
    Open-weight large language model (LLM) zoos allow users to quickly integrate state-of-the-art models into systems. 
    Despite increasing availability, selecting the most appropriate model for a given task still largely relies on public benchmark leaderboards and educated guesses.
    This can be unsatisfactory for both inference service providers and end users, where the providers usually prioritize cost efficiency, while the end users usually prioritize model output quality for their inference requests.
    In commercial settings, these two priorities are often brought together in Service Level Agreements (SLA). 
    We present \algname, an online stochastic optimization algorithm for energy-optimal model selection from a model zoo, which works on a per-inference-request basis. 
    For a given SLA that requires high accuracy, we are up to $2.5\times$ more energy efficient with \algname than with randomly selecting an LLM from the zoo while maintaining SLA quality constraints.
\end{abstract}

\section{Introduction}
\label{sec:intro}
\input{chapters/introduction}

\section{MESS+: Model Selection With Energy-Optimal Service Level Guarantees}
\label{sec:methodology}

\input{chapters/methodology}

\vspace{-0.5\baselineskip}
\section{Experiments}
\vspace{-0.5\baselineskip}
\label{sec:experiments}

\input{chapters/experiments}

\vspace{-0.5\baselineskip}
\section{Conclusions}
\vspace{-0.5\baselineskip}
\label{sec:conclusions}
\input{chapters/conclusions}

\section*{Acknowledgements}
This work is partially funded by the Bavarian Ministry of Economic Affairs, Regional Development and Energy (Grant: DIK0446/01). 
We would like to thank the PANDORA project (\url{https://pandora-heu.eu/}) for our fruitful discussions.

{
    \bibliographystyle{plainnat}
    \bibliography{main}
}

\newpage
\appendix

\section*{Appendix}
\label{sec:appendix}
\input{chapters/appendix}

\end{document}

%% file: chapters/introduction.tex
As the number of open-weight large language models (LLMs), such as Llama~\cite{llama3_paper}, Mistral/Mixtral~\cite{jiang2024mixtralexperts}, and Granite~\cite{granitegranite}, is increasing rapidly, deep learning infrastructure providers and end users are confronted with an abundance of models (model zoo) for their language modeling tasks. 
This leaves many users questioning what model is best to choose and whether highly regarded benchmark results apply to their specific problem~\cite{Humza2023}.
Currently, the best way to approach model selection is educated guessing.
Since working with LLMs can be expensive~\cite{Samsi2023}, minimizing costs is an equally high priority for end users and inference endpoint operators. This leaves us with a tri-fold problem:

\textbf{End-users primarily care about a correct model output}.
When inquiring about text information, e.g., by asking questions or requesting language translation, end users are mostly interested in obtaining factually correct and sufficiently clear language output~\cite{pmlr-v235-wettig24a}.
Additionally, many users are unfamiliar with the technical details of LLMs, making it challenging for them to select the right model for the job, i.e., their primary references are domain-specific benchmark rankings \cite{open-llm-leaderboard-v2, eval-harness}. 
However, there is no intuitive method to compare the complexity of individual requests with benchmark tasks. 
Thus, we require an automatic method to select the most appropriate LLM for any given request. 

\textbf{Inference endpoint providers prioritize low operating costs}. 
Operating infrastructure that can run state-of-the-art LLMs can be costly. 
Microsoft has announced it will acquire a stake in the Three Mile Island nuclear power plant in the United States to satisfy the energy demand of its planned data center in Pennsylvania, which has two reasons: consistent energy delivery and low energy cost \cite{reuters2024}.
In times of globally rising energy costs, this underpins the necessity of energy-optimal service operations. 
Currently, inference service providers only allow their users to query specific models on serverless endpoints or to deploy individual models on dedicated hardware. 
To further optimize operating costs and improve user experience, we require a method that can choose the best model for any given request while minimizing energy consumption.

\textbf{Enterprise use-cases require a consistently high-quality model inference output while keeping costs in check}.
Enterprise users unite the requirement for high-quality model outputs and price sensitivity.
Thus, commercial players typically rely on service-level agreements (SLAs) when sourcing services for their own products. 
This creates a legal basis for holding the model operator responsible for delivering high quality and performance.
Such SLAs typically come with various levels where, in the case of model outputs, the primary quality metric is accuracy. 
Thus, we require a method for formalizing and quantifying the SLA requirements.

In summary, we ask the question: 
\vspace{-6pt}
\begin{quote}
    \textit{Can we select appropriate models from the model zoo to ensure energy efficiency while satisfying SLAs?}
\end{quote}
\vspace{-6pt}
In this paper, we address this question by using a stochastic optimization~\cite{neely2022stochastic} framework to develop an optimal control algorithm, which enables end users to query an inference service and automatically select the most appropriate \textbf{M}odel with \textbf{E}nergy-optimal \textbf{S}ervice-level Guarantee\textbf{S} (\algname).

Our work is related to two major research directions: dynamic inference and inference request scheduling. A broad overview is provided in \cite{zhou2024_survey}.

\textbf{Dynamic inference}. 
Typically, dynamic inference approaches involve early exit strategies within a single model and require changes to the original model architecture. 
These changes are usually additional layers that decide whether a request continues to subsequent layers or is evicted \cite{liu-etal-2020-fastbert,xin-etal-2020-deebert}. 
While these approaches can save inference costs, they require additional training of the decision layers. 
Further, increasing the capabilities of dynamic inference models is difficult as it requires changing the model architecture and re-training. 
\algname removes the need for altering the architecture of readily available pre-trained models and automatically selects the most appropriate model with regard to a given SLA for each request.

\textbf{Inference request scheduling}.
To this end, scheduling work primarily focuses on reducing latency during an inference call. 
Here, the main idea is to group \cite{kwon2023_vllm}, arrange \cite{holmes2024_ds}, or early evict \cite{wu2023_inference} inference requests for faster processing. 
These approaches focus exclusively on scenarios with a single fixed model for all incoming requests and a priority for latency.
In contrast, \algname is a decision-making method to route requests for energy-optimal processing while maintaining a minimum accuracy over time. 
As such, our approach contributes to establishing minimum quality guarantees but does not consider latency directly. 
However, there is a relationship between latency and energy efficiency since smaller models are typically faster to execute an inference call \cite{Samsi2023}.
Since \algname routes inference requests to different models, it can build on top of existing scheduling techniques.

Taken together, we enable dynamic inference across readily available pre-trained LLMs with service level guarantees, while offering compatibility with existing inference load scheduling techniques. \algname can reduce the energy consumption of a given task by up to $4.6\times$ compared to a fixed selection of a single model from a model zoo.

%% file: chapters/methodology.tex
The overall goal of \algname is to find the most suitable LLM for each inference request $t \in T$ to minimize the energy consumption $E_m(t)$ under model performance constraints defined by an SLA over time to ensure contractual compliance.

\subsection{Overall Problem Formulation}
We first formulate the optimization problem and introduce typical constraints that come with SLAs.

\textbf{Inference Cost} (Objective). 
The energy costs for querying a model can vary significantly over time and depend on the chosen model, i.e., our computational costs are volatile in model zoos.
For instance, a model zoo can include an LLM with 1B parameters and another with 13B parameters. The smaller LLM requires significantly fewer resources than the larger model.

\textbf{Service-Level Agreement} (Constraints). 
In SLAs, we typically find a minimum service quality requirement. We consider a minimum average accuracy over requests, denoted by $\alpha.$\footnote{In practice, $\alpha$ should be chosen with a certain safety margin from the SLA requirement such that we do not violate the SLA even if the average accuracy is slightly below $\alpha$.
} We denote each model's accuracy for processing the inference request $t$ as $A_m(t)$, and we choose exactly one model for any $t$ to identify the most appropriate model.

\textbf{Overall control problem}. 
Taken together, the objective and constraints can be formalized into the following problem of minimizing the average energy consumption per request under performance constraints defined in an SLA:
\begin{subequations}
\begin{align}
    \textstyle \min_{\{y_m(t):\forall t, m\}} \quad & \textstyle\frac{1}{T} \sum_{t=1}^T \sum_{m=1}^M y_m(t) E_m(t), \label{eq:energy_minimization} \\
    \text{s.t.} \quad & \textstyle\frac{1}{T} \sum_{t=1}^T \sum_{m=1}^M y_m(t) A_m(t) \geq \alpha, \label{eq:accuracy_constraint} \\
     & \textstyle\sum_{m=1}^M y_m(t) = 1, \forall t \in \{1, \dots, T\}, \label{eq:one_model_constraint} \\
     & y_m(t) \in \{0, 1\}, \forall t, m, \label{eq:binary_constraint}
\end{align}
\end{subequations}
where $y_m(t)=1$ if model $m$ is chosen and $y_m(t)=0$ otherwise.

\textbf{Challenges}. 
Our optimization problem involves an inherent trade-off between model accuracy and energy consumption, since larger LLMs, and thus more capable ones, typically yield higher accuracy while consuming more energy at the same time.
As such, we see a \textit{correlation over time between} the objective and constraints.    
Further, optimizing the energy costs involves a time average that is hard to estimate a priori as the properties of future requests are generally \textit{unknown} and \textit{heterogeneous}.
Similarly, $A_m(t)$ is only available \textit{after} querying the model and only if there is a \textit{feedback signal} on whether the response is satisfactory or not.

\subsection{Translation into an Online Decision Making Problem}
To address the aforementioned challenges, we introduce an online decision-making process. 
Here, the quantities of $E_m(t)$ and $A_m(t)$ are captured in every request, i.e., for every arriving $t$, without knowledge of future statistics. 

\textbf{Methodology}. We base our approach on the Lyapunov drift-plus-penalty framework \cite{neely2022stochastic}. 
Since SLAs are often volume-based, service quality guarantees are typically given for a maximum number of requests or transactions. 
Thus, we deviate from \cite{neely2022stochastic} and assume a finite $T$ instead of an infinite $T$.

\textbf{Virtual Queues}. We use a virtual queue with length $Q$ to capture SLA violations,
where 
\begin{align}
    \scriptsize
    \textstyle Q(t+1) \leftarrow \max \left\{0, Q(t) + \alpha - \sum_{m=1}^M y_m(t)A_m(t)\right\},
    \label{eq:queue-update}
\end{align}
where for $t=1$, we initialize $Q(1)$ to either zero or a small positive value.
Intuitively, this captures the cumulative violations. 
Hence, we aim to collectively minimize our objective \eqref{eq:energy_minimization} and the queue length.  We note that the average accuracy over requests needs to be greater than or equal to $\alpha$ in the constraint, while the direction of inequalities in the constraint is opposite in \cite{neely2022stochastic}, thus our queue update equation in \eqref{eq:queue-update} is slightly different from that in \cite{neely2022stochastic}.
    
\textbf{Decision problem for each request}. 
For each inference request $t$, we aim to minimize energy consumption while complying with SLA requirements. We formulate this trade-off by
\begin{subequations}
\begin{align}
    \textstyle \min_{\{y_m(t):\forall m\}} \quad & \textstyle V\cdot \sum_{m=1}^M y_m(t) E_m(t) + Q(t) \left(\alpha - \sum_{m=1}^M y_m(t)\hat{A}_m(t)\right)\mathrm{,} \label{eq:per-request_obj} \\
    \textrm{s.t.}\quad & \textrm{Constraints \eqref{eq:one_model_constraint}, \eqref{eq:binary_constraint}},
\end{align}%
\label{eq:per-request-optimization}%
\end{subequations}%
where $\hat{A}_m(t)$ is a predictor of the \textit{estimated output} of $A_m(t)$ because it is impossible to know the exact accuracy before the LLM processes the request.\footnote{Note that, in this work, we assume that the accuracy can be estimated immediately after the LLM generates its output. Therefore, in \eqref{eq:queue-update}, we use $A_m(t)$ to update the virtual queue length, but we use $\hat{A}_m(t), \forall m$, to solve the per-request optimization problem \eqref{eq:per-request-optimization}. The more realistic scenario where the accuracy is not known even after obtaining the LLM output is left for future work.} We will describe the computation of $\hat{A}_m(t)$ in Section~\ref{subsec:accuracy_prediction}.
We consider the energy consumption per request to be known, i.e., we measured the cost for an inference call before adding a model to the model zoo, which can depend on some characteristics of the input request such as its length.
As SLAs come in various configurations, we introduce the control parameter $V > 0$ that caters to different quality requirements. A high value of $V$ increases the priority of energy efficiency and lowers the need for accuracy. 

\textbf{Solution to \eqref{eq:per-request-optimization}}.
Due to Constraints \eqref{eq:one_model_constraint}, \eqref{eq:binary_constraint}, the problem in \eqref{eq:per-request-optimization} can be easily solved using a linear search by setting $y_m(t)=1$ and $y_{m'}(t)=0$ (for all $m'\neq m)$, for each $m\in\{1,2,\ldots,M\}$, and comparing the values of the objective \eqref{eq:per-request_obj}. This procedure has a linear complexity of $\mathcal{O}(M)$. Using similar proof techniques as those in \cite{neely2022stochastic}, we can show constraint satisfaction guarantee, i.e., SLA guarantee, and optimality as $T\rightarrow\infty$. The detailed proof is left for a future version of this work.

\subsection{Predicting the Accuracy for Each Request}
\label{subsec:accuracy_prediction}

We now describe how to obtain the predicted accuracy $\hat{A}_m(t)$, $\forall m, t$, which is needed to solve \eqref{eq:per-request-optimization}. To facilitate the description, we write out $\hat{A}_m(t) = \hat{A}(\mathbf{x}_{m,t}, \mathbf{a}_t)$, where $\mathbf{x}_{m,t}$ is the parameter vector of the (small) accuracy estimation model for the $m$-th LLM used for request $t$, and $\mathbf{a}_t$ is the $t$-th input request. We omit the subscript $t$ in $\mathbf{x}_{m,t}$ in the following when it is unnecessary to specify the parameter used for a specific request $t$.
We learn $\mathbf{x}_m$ through a probabilistic exploration procedure. To \textit{explore} the model zoo, we query multiple models with the same input request to obtain their actual accuracies $A_m(t)$, allowing us to learn $\mathbf{x}_m$ from $A_m(t)$ over time. 
More specifically, we sample from a distribution $\mathcal{X}_t \sim \mathrm{Bernoulli}(p_t)$, where the exploration probability $p_t \leftarrow \min(1, \frac{c}{\sqrt[3]{t}})$ and  parameter $c>0$ controls the exploration likelihood. 
The larger $c$, the more likely it is to do an exploration step.
We decay the probability $p_t$ over time as the estimation $\hat{A}_m(t)$ improves with each exploration step.

\begin{wrapfigure}[22]{R}{0.55\textwidth}
    \vspace{-2\baselineskip}
    \begin{algorithm}[H]
        \caption{Selecting the \textbf{M}odel with \textbf{E}nergy-optimal \textbf{S}ervice level Guarantee\textbf{S} (\algname)
        }
        \label{algo:messplus}
        \SetCustomAlgoRuledWidth{0.45\textwidth}
        \scriptsize
        \KwInput{
            $T$; 
            $V$;
            $\alpha$;
            $c$;
            $\{E_m(t): \forall m, t\}$;
            learning rate $\eta>0$
        }
        
        \KwOutput{Outputs of models chosen for all $t$}
        
        Initialize $Q(1) \leftarrow 0$; predictor parameters $\mathbf{x}_m$ to a common random vector for all $m$; $k\leftarrow 1$ \;
        
        \For{$t \leftarrow 1$ \KwTo $T$}{
            Compute $p_t \leftarrow \min\left(1, \dfrac{c}{\sqrt[3]{t}}\right)$; \\ Sample $\mathcal{X}_t \sim \text{Bernoulli}(p_t)$\;
            
            \eIf{$\mathcal{X}_t = 1$}{
                \tcp{Exploration}
                \ForEach{$m \in \{1, 2, ..., M\}$}{
                    Obtain true accuracy $A_m(t)$; 
                    
                    $\mathbf{x}_{m, t+1} \leftarrow \mathbf{x}_{m, t} - \eta\nabla_{\mathbf{x}}\left( \hat{A}(\mathbf{x}_{m,t}, \mathbf{a}_t) - A_m(t) \right)^2$\;
                }
                $m^{*}\leftarrow\arg\max_m A_m(t) $\;
            }{
                \tcp{Solve \eqref{eq:per-request-optimization}}
                $m^* \leftarrow\arg\min_m V \cdot E_m(t) + Q(t) \cdot (\alpha - \hat{A}_m(t))$\;
                $\mathbf{x}_{m, t+1} \leftarrow \mathbf{x}_{m, t}$\;
            }
        Get output from model $m^{*}$ and its accuracy $A_{m^{\!*}}(t)$\;
        \tcp{Virtual queue update} 
        $Q(t+1) \leftarrow \max\{0, Q(t) + \alpha - A_{m^{\!*}}(t)\}$; 
        }
    \end{algorithm}
\end{wrapfigure}

Especially for the first set of arriving requests, when we do not know how to choose the optimal LLM for request $t$, we must explore each $m$ in the model zoo for the best-performing model for $t$. 
While doing so, we capture the actual model accuracy $A_m(t)$ of each $m$, which we use to learn $\mathbf{x}_m$.
We define the mean square error (MSE) objective of accuracy prediction for an estimation over all the possible incoming requests: 
\begin{equation}
    \textstyle L(\mathbf{x}_{m}) = \mathbb{E}_{\mathbf{a}_t} \left( \hat{A}(\mathbf{x}_{m}, \mathbf{a}_t) - A_m(t) \right)^2\mathrm{.}
    \label{eq:accuracy-predictor-loss}
\end{equation}
If we learn $\mathbf{x}_{m}$ using stochastic gradient descent (SGD), assuming that the distribution of the input request $\mathbf{a}_t$ is IID across $t$, it is easy to prove that the convergence upper bound is
\begin{equation}
    \textstyle \frac{1}{K}\sum_{k=1}^K \mathbb{E}\|\nabla L(\mathbf{x}_{m, t_k})  \|^2 \leq \mathcal{O}\big(\frac{1}{\sqrt{K}}\big),
    \label{eq:sgd-bound}
\end{equation}
where $K$ is the number of exploration steps and $t_k$ is the request index corresponding to the $k$-th exploration step. 

When exploring, we always use the output from the largest model as the final model output as we have already spent the energy to query the largest model, i.e., we \textit{do not} use the solution from \eqref{eq:per-request-optimization} in this case.
Since we query all the models in an exploration step, it comes at additional energy costs.
We can upper bound the \textit{average} \textit{additional} energy consumption per exploration step over time by 
\begin{equation}
    \textstyle \frac{1}{T}\sum_{k=1}^K E =\frac{KE}{T}, 
    \label{eq:energy-bound}
\end{equation}
where $E$ is the maximum energy consumption for querying the entire model zoo.

By choosing $p_t$ as described above, we have $\mathbb{E}[K]=\Theta\big(T^{\frac{2}{3}}\big)$. Then, the SGD convergence bound in \eqref{eq:sgd-bound} becomes $\mathcal{O}\left(\sfrac{1}{\sqrt[3]{T}}\right)$ and the upper bound of average additional energy consumption due to exploration in \eqref{eq:energy-bound} becomes
$\mathcal{O}\left(\sfrac{E}{\sqrt[3]{T}}\right)$. Both of them approach zero as $T\rightarrow \infty$.
We leave the formal statement and full proof to a future version of this work.

\textbf{Full algorithm}.
The full procedure is described in Algorithm~\ref{algo:messplus}.

%% file: chapters/experiments.tex
We demonstrate the effectiveness of \algname with state-of-the-art language modeling tasks, including WMT14 \cite{bojar2014findings} for language translation and CNNDailyMail \cite{see-etal-2017-get} for text summarization. 
We construct our model zoo from the TinyLlama~1.1B~\cite{zhang2024tinyllama} and Llama-2 13B~\cite{touvron2023llama2} models, each representing a different model class based on its number of model parameters.
To measure the performance of \algname in choosing the most appropriate model in the model zoo, we use the \mbox{BLEU-1}~\cite{papineni2002bleu} metric to evaluate the language translation task and the ROUGE-1 score \cite{lin2004rouge} for the summarization task. 
We establish one SLA for each task, where the requirement is to reach an average BLEU score of $ \alpha = 0.52$ and ROUGE score of $\alpha = 0.315$ for the translation and summarization tasks, respectively.
We provide additional experimental details in \Cref{sec:additional_experimental_details}.

Since $V$ and $c$ have to be chosen manually, we run a set of ablation studies to explore the sensitivity of both parameters. 
Varying the importance of energy efficiency $V$ between $[0.01, 100]$ for both tasks while complying with their respective SLA shows that the maximum $V$ we can choose to reliably achieve $\alpha$ is $V = 0.1$. 
For $V > 0.1$, we violate the SLA.
Interestingly, when looking at values of $[1, 10]$ for $c$, which controls the exploration likelihood, we observe that with $c = 3$, $\hat{A}(t)$ yields the lowest loss over time and thus the best performance.
Hence, we choose $V = 0.1$ and $c=3$ for our experiments. Further details on selecting $V$ and $c$ are  in Appendices~\ref{sec:v_sensitivity} and \ref{sec:c_sensitivity}, respectively.

We look at four different scenarios: (I) choosing the smallest model only (TinyLlama-1.1B), (II) choosing the largest model only (Llama-2 13B), (III) choosing a model randomly while satisfying the constraints, and (IV) applying \algname.
The results are shown in \Cref{tab:baseline_comparison}. In the case of (I), we do not satisfy the SLA in either task, while for (II), we observe the highest accuracy and comply with the SLAs, but the energy consumption is also the highest.
Case (III) also complies with the SLAs and reduces the energy consumption by $1.9\times$ and $1.8\times$ for the translation and summarization tasks, respectively, when compared to (II).
However, (IV) provides the same average accuracy over time as (III) while reducing the energy consumption by $3.5\times$ and $4.6\times$ for the translation and summarization tasks, respectively, when compared to (II).
Thus, we see that \algname strictly ensures SLA compliance and reliably reduces energy consumption compared to randomly choosing an available model that satisfies the SLA requirements. Since energy costs are directly proportional to energy usage, the energy savings brought by \algname are directly related to cost reduction. With this, \algname can help reduce the operating costs for model serving. \vspace{-1em}

\begin{table}[H] 
    \centering
    \caption{Comparison of baseline model selection strategies and \algname with  $V=0.1$ and $c=3$. Our approach satisfies the SLA with requirement $\alpha$ while consuming the least energy among all compliant strategies. 
    }
    \vspace{0.3\baselineskip}
    \label{tab:mess_performance}
    \resizebox{0.85\textwidth}{!}{
        \begin{tabular}{lrrcrrc}
            \toprule
            \textbf{Model} & \multicolumn{3}{c}{\textbf{WMT14 ($\alpha = 0.52$)}} & \multicolumn{3}{c}{\textbf{CNNDailyMail ($\alpha = 0.315$)}} \\
            \cmidrule(lr){2-4} \cmidrule(lr){5-7}
            & \makecell{\textbf{Accuracy} \\ (BLEU)} & \makecell{\textbf{Energy} \\ (Joules)} & \textbf{Meets $\alpha$} & \makecell{\textbf{Accuracy} \\ (ROUGE1)} & \makecell{\textbf{Energy} \\ (Joules)} & \textbf{Meets $\alpha$} \\
            \midrule
            TinyLlama & {\red{49.1}} $\pm$ 0.6 & 44.639 $\pm$ 0.6 & \red{No} & {\red{30.9}} $\pm$ 0.3 & 142.080 $\pm$ 1.4 & \red{No} \\
            Llama-2 13B & {\green{55.1}} $\pm$ 0.4 & 527.870 $\pm$ 5.1 & \green{Yes} & {\green{32.2}} $\pm$ 0.3 & 750.285 $\pm$ 7.5 & \green{Yes} \\
            Random with constraint & {\green{52.0}} $\pm$ 0.0 & 280.426 $\pm$ 3.0 & \green{Yes} & {\green{31.5}} $\pm$ 0.0 & 416.466 $\pm$ 4.3 & \green{Yes} \\
            \algname ($V=0.1$, $c=3$) & {\green{52.2}} $\pm$ 0.2 & \textbf{149.399} $\pm$ 1.3 & \green{Yes} & {\green{31.5}} $\pm$ 0.1 & \textbf{163.836} $\pm$ 1.5 & \green{Yes} \\
            \bottomrule
        \end{tabular}
    }
     \vspace{-1.5em}
    \label{tab:baseline_comparison}
\end{table}

%% file: chapters/conclusions.tex
We present \algname, a novel method for automatic model selection in model zoos for language inference services on a per-request basis. 
The approach is particularly useful for inference providers that cater to users with quality-sensitive workloads as it enables service level guarantees.
The experimental evaluation of \algname demonstrates the effectiveness in routing requests to the most appropriate model while achieving a minimum required accuracy and reducing energy consumption at the same time.
This will lead to overall lower operating costs for inference service providers.

In future work, we will simplify the approximation mechanism that yields $\hat{A}_m(t)$ by removing the need for multiple linear classifiers as this would otherwise establish a new bottleneck with growing model zoos.

%% file: chapters/appendix.tex
\section{Additional Experimental Details}
\label{sec:additional_experimental_details}

\textbf{Benchmark datasets}. 
We use the WMT14 dataset \cite{bojar2014findings} to evaluate how well our approach works with language translation and the CNNDailyMail \cite{see-etal-2017-get} dataset for sequence-to-sequence modeling, specifically highlighting summarization from news articles.

\textbf{Metrics}. 
Since the users sending inference requests are primarily interested in retrieving high-quality results, we use the BLEU-1 \cite{papineni2002bleu} metric to evaluate the language translation task and the ROUGE-1 score \cite{lin2004rouge} for the sequence-to-sequence task. 
To account for the service provider's cost-saving priority, we measure the energy consumption in total joules consumed. 

\textbf{Inference models}. 
We populate $M$ with  TinyLlama~1.1B~\cite{zhang2024tinyllama} and Llama-2 13B~\cite{touvron2023llama2}. 
These models have shown strong performance across language modeling tasks and represent two different model classes: small and large.

\textbf{\algname accuracy predictor}. 
In our experiment, we use one linear model per inference model built with the \texttt{fasttext} library to learn \(\mathbf{x}_m\)~\cite{bojanowski2017enriching,joulin2017bag}. 
We query each linear model for every request \(t\) to estimate the corresponding inference model performance.

\textbf{Application scenarios}.
To demonstrate the sensitivity of control parameter $V$ in \algname, we explore different service levels as they would be found in commercial services. 
For the WMT14 translation task, users can choose from three different plans: \textit{silver}, \textit{gold}, and \textit{platinum}, where the target accuracy is $0.49$, $0.5$, and $0.51$, respectively.
To discuss the energy efficiency implications, we aim to achieve $\alpha = 0.52$ for WMT14 and $\alpha = 0.315$ for CNNDailyMail. 
Both scores are considered high \cite{lin2004rouge,papineni2002bleu}.

\section{Additional Experimental Results}
In this section, we discuss the impact of our control parameters $V$ (the priority of energy efficiency) and $c$ (the likelihood for exploration steps).

\subsection{Varying the Priority for Energy Efficiency}
\label{sec:v_sensitivity}
\begin{figure}[t]
    \centering
    \includegraphics[width=1\textwidth]{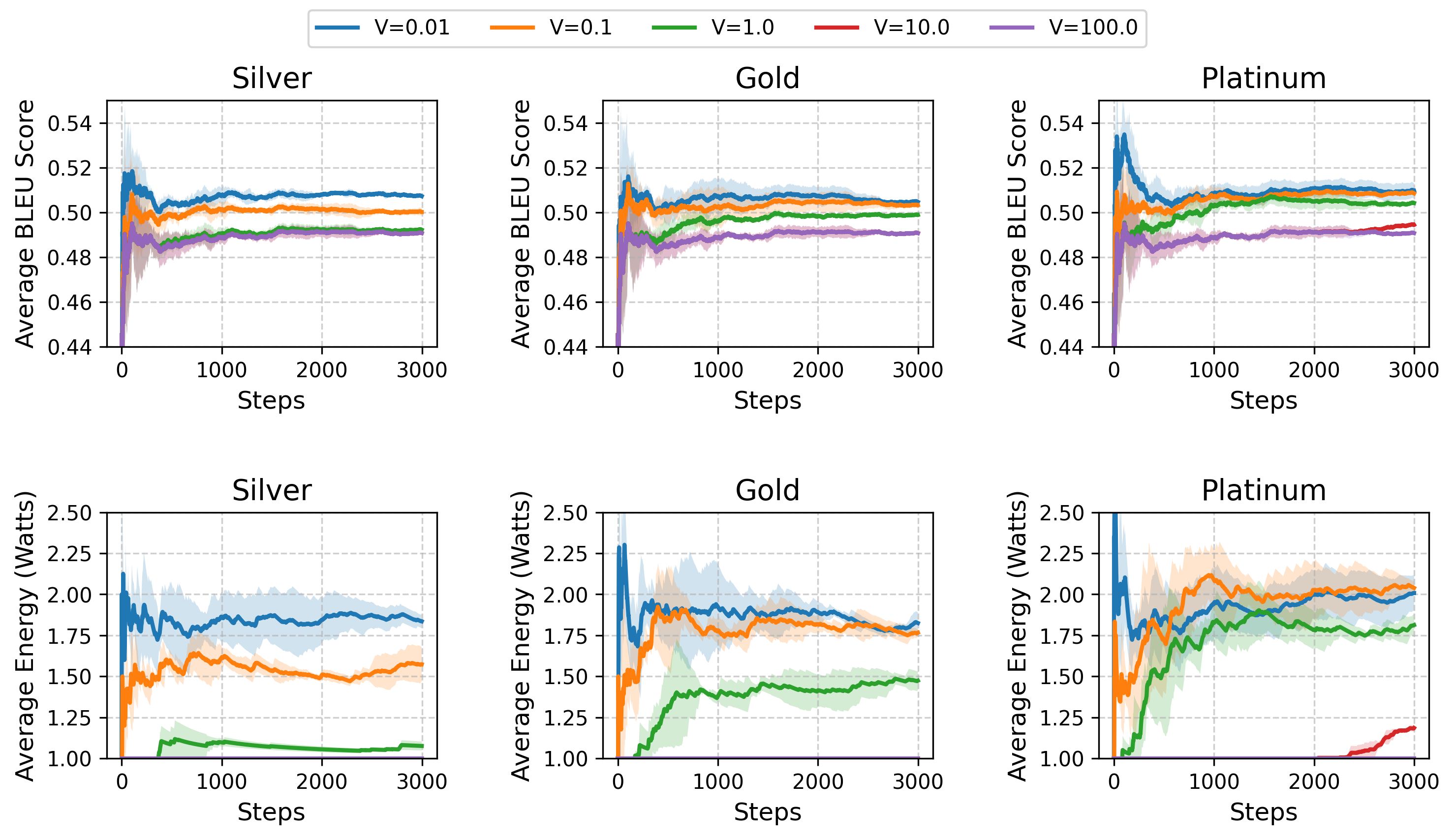}
    \caption{Service Level Guarantees and energy consumption with varying $V$ across different minimum accuracy values $\alpha$.}
    \label{fig:bleu_energy_comparison}
\end{figure}

The parameter \(V\) serves as a tuning knob that adjusts the trade-off between prioritizing accuracy and minimizing energy usage. By systematically adjusting \(V\), we observe how the system behavior changes in response to the three service level plans. The results are shown in \Cref{fig:bleu_energy_comparison}.

As mentioned above, we introduce three different SLAs that each require a different $\alpha$.  
Between the three $\alpha$ values, we vary the accuracy by exactly 1\% to understand how much more energy we have to use in order to serve the next higher SLA.

For the \textit{silver} level, where the minimum accuracy requirement is low, the system is less sensitive to variations of \(V\). As we increase \(V\) from 0.01 to 100, the average BLEU score remains consistently above the silver threshold. In such a scenario, it is beneficial to set $V$ to a high value as this prioritizes energy efficiency, which will route more requests to the smaller and faster model.

For the \textit{gold} level, the sensitivity to \(V\) becomes more notable. if \(V\) is set too high (e.g. $V>5$, the system excessively prioritizes energy savings, leading to a noticeable drop in the average BLEU score below the acceptable service requirement.

At the \textit{platinum} level, the sensitivity of $V$ increases further compared to gold since the quality requirements have increased as well. However, the viable configurations must now emphasize model accuracy, which leads to $V < 1$. 
Here, a 1\% performance gain costs $1.9\times$ more energy than the gold plan. 

Analyzing \(V\) reveals that its optimal value is inversely related to the minimum service requirement $\alpha$. As $\alpha$ increases, the range of acceptable \(V\) narrows, and the system demands a greater emphasis on accuracy rather than cost reduction. Service providers can leverage this knowledge to adjust \(V\) dynamically based on operational priorities, such as reducing energy costs during non-peak hours and maximizing accuracy during critical periods. 

Generally, for our experimental setup, we find that $V = 0.1$ ensures SLA compliance over time while also minimizing energy consumption.

\begin{figure}
    \centering
    \begin{subfigure}{.35\textwidth}
        \centering
        \includegraphics[width=\linewidth]{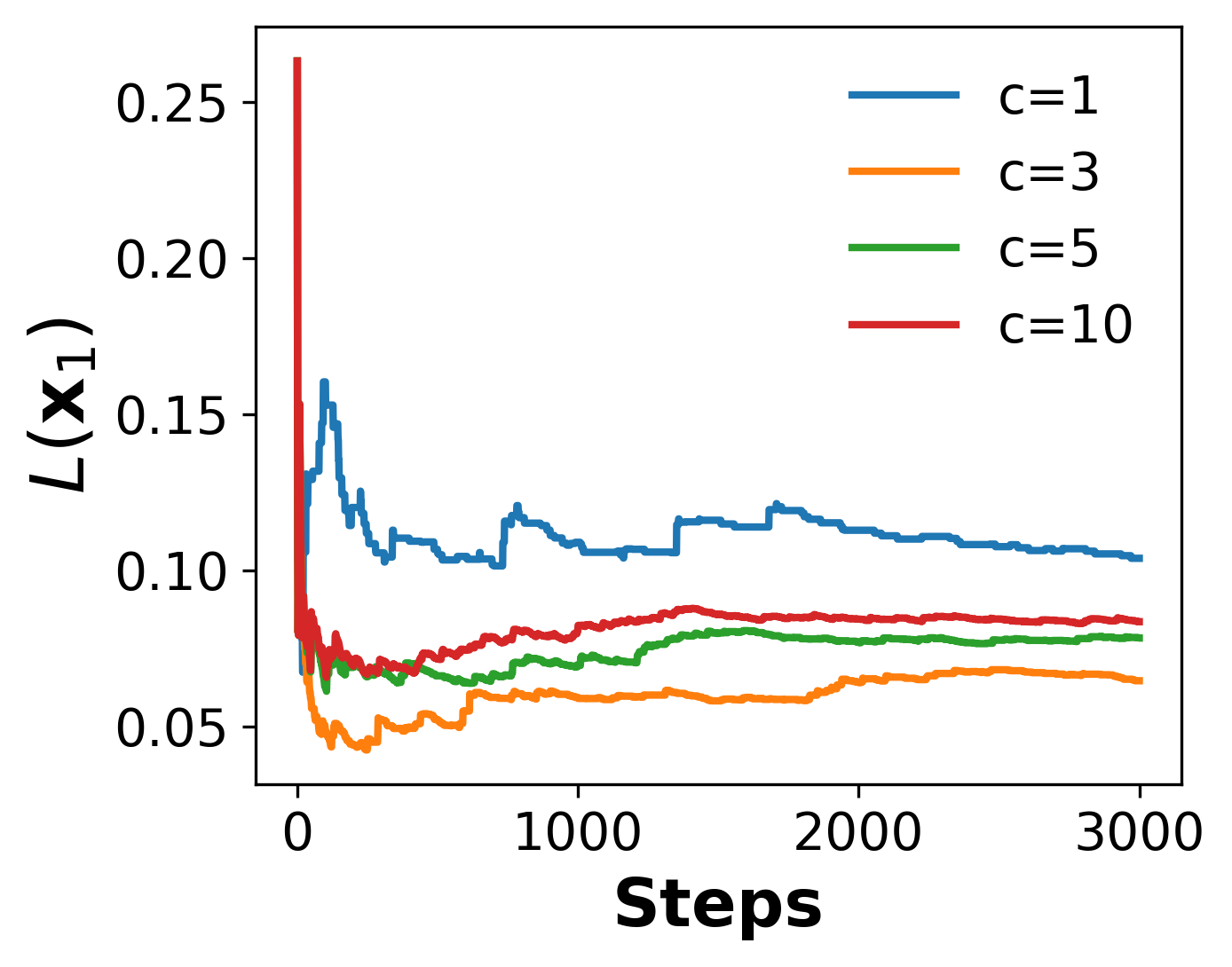}
        \caption{Large model} 
        \label{fig:l_loss_comparison}
    \end{subfigure}
    ~~~~
    \begin{subfigure}{.35\textwidth}
        \centering
        \includegraphics[width=\linewidth]{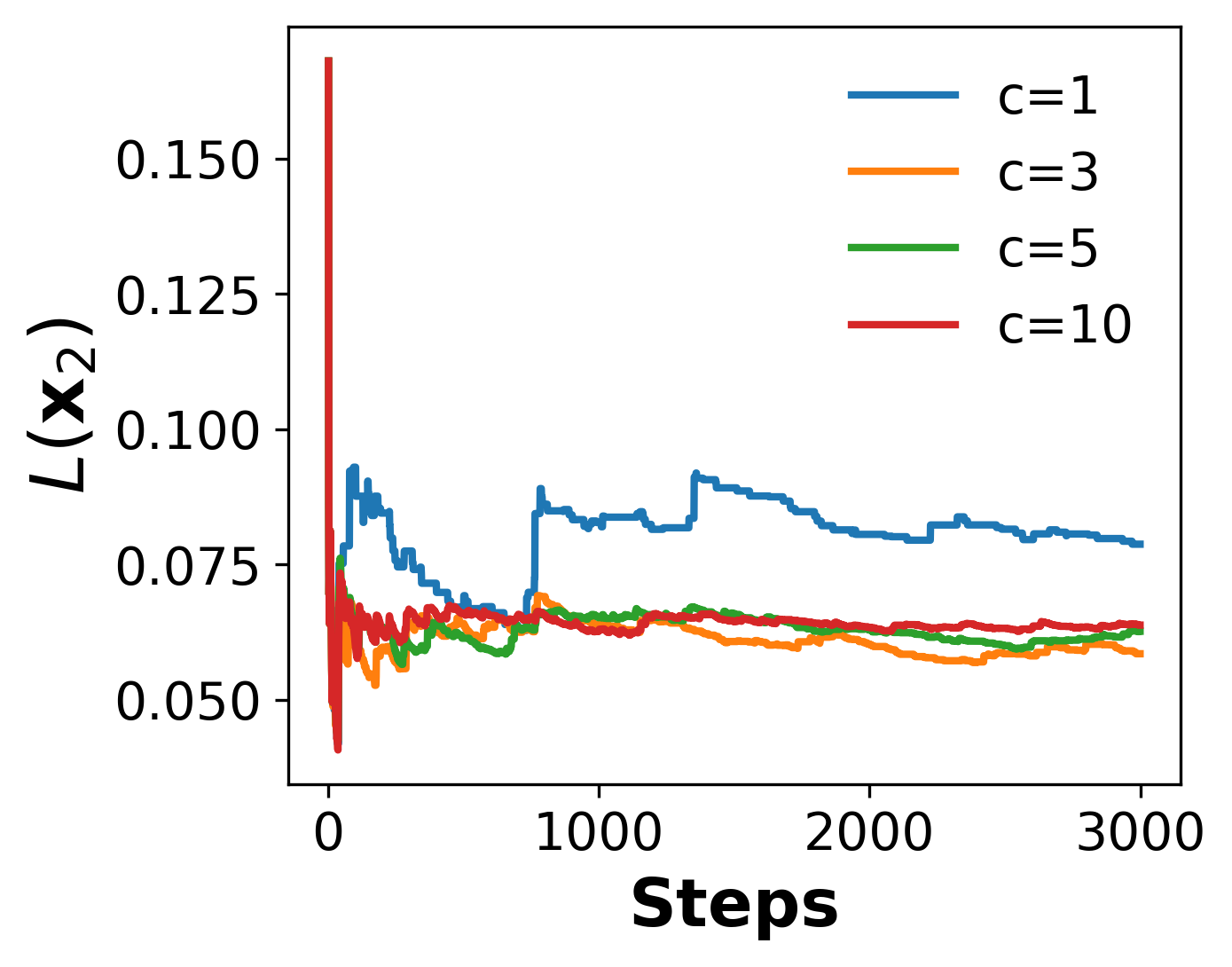}
        \caption{Small model} 
        \label{fig:s_loss_comparison}
    \end{subfigure}
    
    \caption{We observe that $L(\mathbf{x}_m)$ converges to the lowest loss with $c = 3$. 
    }
    \label{fig:c_sensitivity}
\end{figure}

\subsection{Exploring the Likelihood of Exploration in \algname}
\label{sec:c_sensitivity}

With parameter $c$, we control the exploration likelihood over time. 
The higher we set $c$, the higher the likelihood of exploring the model zoo, since $p_t = \min (1, \frac{c}{\sqrt[3]{t}})$, and the number of training steps \(K\) becomes larger. 
However, we decay $p_t$ over time such that we increasingly rely on the model with parameter $\mathbf{x}_{m,t}$ to predict the accuracies of different models for each request $t$. 

We explore the effects of varying levels of $c$ ranging from $[1, 10]$ on the loss of the accuracy predictors, defined in \eqref{eq:accuracy-predictor-loss}, for different models. 
This is an indicator of how well $\hat{A}_m(t)$ approximates $A_m(t)$. The results are shown in \Cref{fig:c_sensitivity}.

We find that balancing the level of exploration is important since $c = 1$ leads to less training; therefore, the training samples to learn $\mathbf{x}_m$ are significantly fewer than with $c = 10$, for instance. 
A large $c$ implies more training of $\mathbf{x}_m$ but can also lead to overfitting on the incoming requests.
While the convergence speed is similar for all values of $c$, we find that $c = 3$ leads to the lowest loss. 
Therefore, our systematic parameter search for $c$ yields the best accuracy prediction performance with $c = 3$. 
We need between $200$ and $250$ exploration steps for the classifier to fully converge. 
We also see that over time, each classifier overfits the data, which explains the need for decreasing $p_t$ over time.

The consequence of increased latency is observed as we increase \(c\). As shown in \Cref{tab:c_configurations_time}, the average time per inference request increases as $c$ increases.
While \algname aims to establish minimum service guarantees and optimize energy consumption, it is important to consider the trade-off with latency. When \(c = 10\), multiple models are queried more often compared to \(c = 1\). The additional processing introduces greater latency, as querying all \(M\) and training \(\mathbf{x}_m\) requires more time and resources. Increased latency can negatively impact user experience, especially in real-time services where prompt response time is crucial. 

\begin{table}[H]
    \centering
    \caption{More exploration leads to a longer request processing latency. We evaluate the effects of choosing different $c$ values on the average inference time in seconds.}
    \label{tab:c_configurations_time}
    \begin{tabular}{lrr}
        \toprule
        & \multicolumn{2}{c}{\algname Average Request Processing Duration} \\
        \cmidrule{2-3}
        \textbf{$c$} & \textbf{WMT14} & \textbf{CNNDailyMail} \\
        \midrule
        1 & 0.99s & 1.11s \\
        3 & 2.86s & 3.13s \\
        5 & 4.02s & 4.34s \\
        10 & 5.38s & 5.76s \\
        \bottomrule
    \end{tabular}
\end{table}